\documentclass[runningheads]{llncs}

\usepackage{graphicx}
\usepackage{epsfig}
\usepackage{url}
\urldef{\mailsa}\path|costapan@eng.auth.gr,  petroula@gen.auth.gr|
\begin{document}
\title{The Margitron: A Generalised Perceptron with Margin}
\author{Constantinos Panagiotakopoulos \and Petroula Tsampouka } 


\institute{Physics Division, School of Technology \\ Aristotle University of Thessaloniki, Greece\\
\mailsa\\
}

\maketitle

\begin{abstract} 
We identify the classical Perceptron algorithm with margin as a member of a broader family of large margin classifiers  which we collectively call the Margitron. The Margitron, (despite its) sharing the same update rule with the Perceptron, is shown  in an incremental setting to converge in a finite number of updates to solutions possessing any desirable fraction of the maximum margin. Experiments comparing the Margitron with decomposition SVMs on tasks involving linear kernels and 2-norm soft margin are also reported. 
\end{abstract}

\renewcommand{\vec}[1]{\mbox{\boldmath$#1$}}
\newcommand{\vic}[1]{\mbox{$#1$}}
\newcommand{\shri}[1]{\resizebox{!}{0.14cm}{\mbox{$#1$}}}
\newcommand{\shrr}[1]{\resizebox{!}{0.20cm}{\mbox{$#1$}}}
\newcommand{\shhh}[1]{\resizebox{!}{0.23cm}{\mbox{$#1$}}}
\newcommand{\shr}[1]{\resizebox{!}{0.25cm}{\mbox{$#1$}}}
\newcommand{\srr}[1]{\resizebox{!}{0.32cm}{\mbox{$#1$}}}
\newcommand{\shh}[1]{\resizebox{!}{0.36cm}{\mbox{$#1$}}}
\newcommand{\inc}[1]{\resizebox{!}{0.39cm}{\mbox{$#1$}}}
\newcommand{\voc}[1]{\resizebox{!}{0.42cm}{\mbox{$#1$}}}
\newcommand{\vocc}[1]{\resizebox{!}{0.44cm}{\mbox{$#1$}}}
\newcommand{\incc}[1]{\resizebox{!}{0.46cm}{\mbox{$#1$}}}
\newcommand{\incr}[1]{\resizebox{!}{0.47cm}{\mbox{$#1$}}}
\newcommand{\inrr}[1]{\resizebox{!}{0.48cm}{\mbox{$#1$}}}
\newcommand{\inccr}[1]{\resizebox{!}{0.50cm}{\mbox{$#1$}}}
\newcommand{\iccc}[1]{\resizebox{!}{0.52cm}{\mbox{$#1$}}}
\newcommand{\inccc}[1]{\resizebox{!}{0.53cm}{\mbox{$#1$}}}
\newcommand{\incrr}[1]{\resizebox{!}{0.56cm}{\mbox{$#1$}}}
\newcommand{\inr}[1]{\resizebox{!}{0.58cm}{\mbox{$#1$}}}

\section{Introduction}
It is widely accepted that the larger the margin of the solution hyperplane the greater is the generalisation ability of the learning machine \cite{Vap,SBWA}. The simplest online learning algorithm for binary linear classification, the Perceptron \cite{Ros,Nov}, does not aim at any margin. The problem, instead, of finding the optimal margin hyperplane lies at the core of Support Vector Machines (SVMs) \cite{Vap,CST}. Their efficient implementation, however, is somewhat hindered by the fact that they require solving a quadratic programming problem.

The complications encountered in implementing SVMs has respurred the interest in alternative large margin classifiers many of which are based on the Perceptron algorithm. The oldest such algorithm which appeared long before the advent of SVMs is the standard Perceptron with margin \cite{DH}, a straightforward extension of the Perceptron, which, however, in an incremental setting is known to be able to guarantee achieving only up to $1/2$ of the maximum margin that the dataset possesses \cite{KM,LZHS,TST}. Subsequently, various algorithms succeeded in achieving larger fractions of the maximum margin by employing modified perceptron-like update rules. Such algorithms include ROMMA \cite{LL}, ALMA \cite{Gen}, CRAMMA \cite{TST1} and MICRA \cite{TST2}. A somewhat different approach from the hard margin one adopted by most of the algorithms above was also developed which focuses on the minimisation of the 1-norm soft margin loss through stochastic gradient descent. There is a connection, however, between such algorithms and the Perceptron since their unregularised form with constant learning rate is identical to the Perceptron with margin. Notable representatives of this approach are the pioneer NORMA \cite{KSW} and the very recent Pegasos \cite{SSS}.

A question that arises naturally and which we attempt to answer in the present work is whether it is possible to achieve a guaranteed fraction of the maximum margin larger than $1/2$ while retaining the original perceptron update rule. To this end we construct a whole new family of algorithms at least one member of which has guaranteed convergence in a finite number of steps to a solution hyperplane possessing any desirable fraction of the unknown maximum margin. This family of algorithms in which the classical Perceptron with margin is naturally embedded will be termed the Margitron. Hopefully, the algorithms belonging to the margitron family by virtue of being generalisations of the very successful Perceptron will have a respectable performance in various classification tasks.         

Section 2 contains some preliminaries and the description of the Margitron algorithm. Section 3 is devoted to a theoretical analysis. Section 4 contains our experimental results while Section 5 our conclusions.  

\section{The Margitron Algorithm}

In what follows we assume that we are given a training set 
which either is linearly separable from the beginning or becomes separable by an appropriate 
feature mapping into a space of a higher dimension \cite{Vap,CST}. This higher dimensional feature space in which 
the patterns are linearly separable will be the considered space. By placing all patterns in the same position at a distance $\rho$ in an additional dimension we construct an embedding of our data into 
the so-called augmented space \cite{DH}. The advantage of this embedding is that the linear hypothesis in the augmented space becomes homogeneous. Throughout our discussion a reflection with respect to the origin in the augmented space of the negatively labelled patterns is assumed in order to allow for a uniform treatment of both 
categories of patterns. Also, $R\equiv\displaystyle \max_{k} \left\| \vec{y}_{k} \right\|$, with $\vec{y}_{k}$ the $k^{\rm th}$ augmented pattern. Obviously, $R \ge \rho $. 

The relation characterising optimally correct classification of the training patterns $\vec{y}_{k}$ by a 
weight vector $\vec{u}$ of unit norm in the augmented space is
\begin{equation}
\label{gamma}
\vec{u} \cdot \vec{y}_{k}\ge \gamma_{\rm d}\equiv \displaystyle \max_{\tiny{ \vec{u}^{\prime}:\left\|\vec{u}^{\prime}\right\|}=1}
\displaystyle \min_{i}\left \{\vec{u^{\prime}} \cdot \vec{y}_{i}\right \}  \ \ \ \ \forall k \enspace.
\end{equation}
We shall refer to $\gamma_{\rm d}$ as the maximum directional margin. It coincides with
the maximum margin in the augmented space with respect to hyperplanes passing through
the origin if no reflection is assumed. The directional margin $\gamma_{\rm d}$
and the maximum geometric margin $\gamma$ in the original (non-augmented) feature space satisfy the inequality
\[
1\le {\gamma}/{\gamma_{\rm d}}\le {R}/{\rho}\enspace.
\]
As $\rho \to \infty$, ${R}/{\rho}\to 1$ and from the above inequality $\gamma_{\rm d} \to \gamma$ \cite{TST}.  

In the Margitron algorithm the augmented weight vector $\vec{a}_{t}$ is initially set to zero, i.e. $\vec{a}_{0}=\vec0$, and is updated according to the classical perceptron rule
\begin{equation}
\label{update}
\vec{a}_{t+1}=\vec{a}_{t}+\vec{y}_k
\end{equation}
each time a misclassification condition is satisfied by a training pattern $\vec{y}_{k}$. For the misclassification condition we consider two options. The first is to replace the constant functional margin threshold $b>0$ in the misclassification condition of the classical Perceptron with margin by a term proportional to a power of the number of steps (updates) $t$
\begin{equation}
\label{misclast}
\vec{a}_{t}\cdot\vec{y}_{k}\le b\:t^{1-\epsilon} \enspace, \ \ \ \epsilon >0 \enspace.
\end{equation}
As a second option we employ a margin threshold proportional to a power of the length of the augmented weight vector leading to a misclassification condition
\begin{equation}
\label{misclasl}
\vec{a}_{t}\cdot\vec{y}_{k}\le b \left\|\vec a_t \right\|^{1-\epsilon} \enspace, \ \ \ \epsilon >0 \enspace.
\end{equation}
For $t=0$ in both (\ref{misclast}) and (\ref{misclasl}) the threshold is set to $0$ resulting in the first pattern being always misclassified. The Margitron with misclassification condition given by (\ref{misclast}) will be referred to as the $t$-margitron whereas the version with condition given by (\ref{misclasl}) as the $\ell$-margitron. Setting $\epsilon=1$ in both the $t$- and the $\ell$-margitron we recover the Perceptron with margin. Notice that the introduction of a constant learning rate is pointless since it amounts to a rescaling of $b$.

\vspace{-2pt}
\begin{figure}[h]
\centering
\epsfig{file=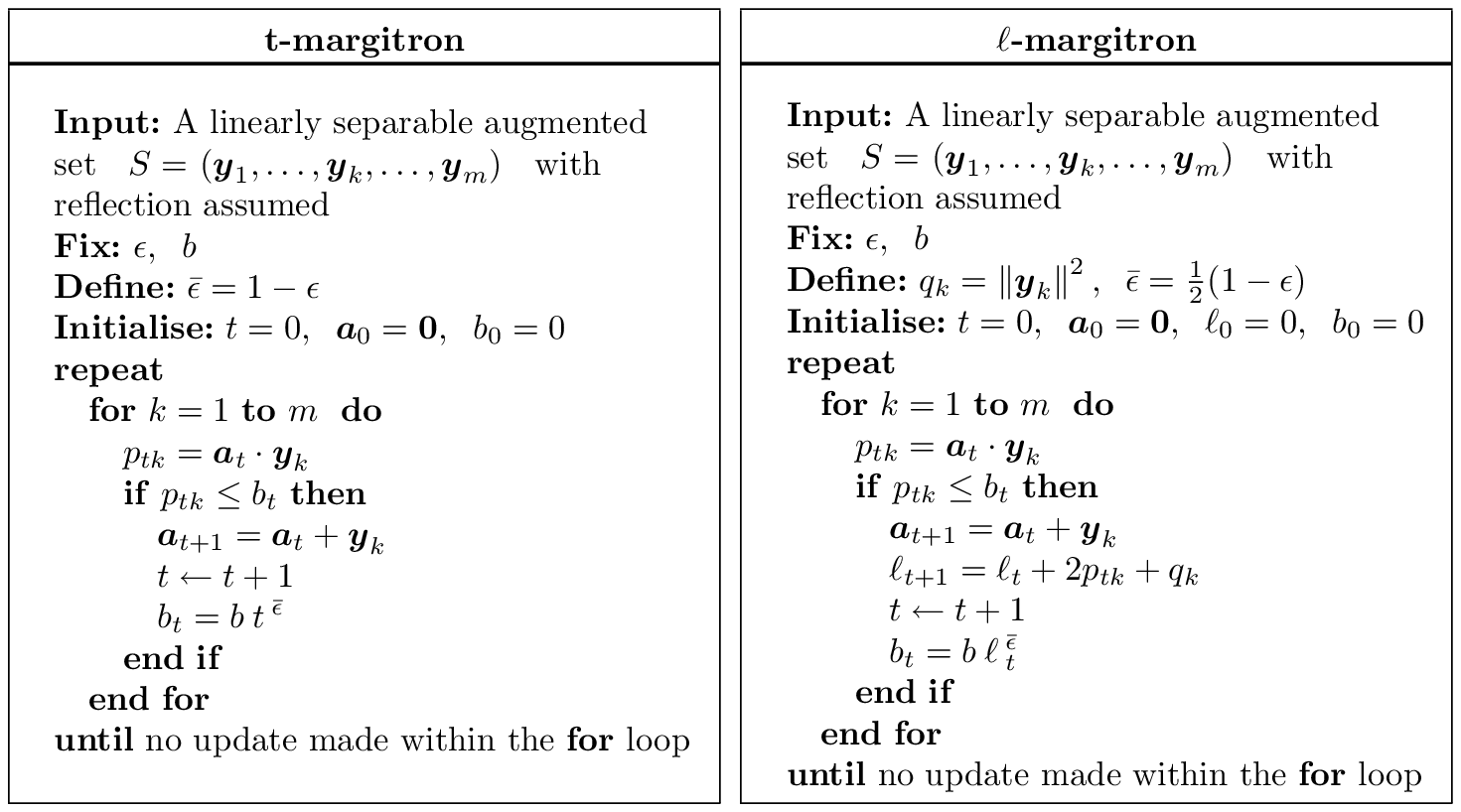, width=1\textwidth}
\caption{The algorithms $t$-margitron and $\ell$-margitron.} 
\end{figure}

Both (\ref{misclast}) and (\ref{misclasl}) can be written for $t>0$ in the form
\begin{equation}
\label{nmisclas}
\vec{u}_{t}\cdot\vec{y}_{k}\le C(t)
\end{equation}
$(\vec{u}_{t}\equiv\vec{a}_{t}/\left\|\vec a_t \right\|, C(t)>0)$ involving the margin $\vec{u}_{t}\cdot\vec{y}_{k}$ in the augmented space of the pattern $\vec{y}_{k}$ with respect to the zero-threshold hyperplane normal to $\vec{a}_{t}$ (i.e. the directional margin of $\vec{y}_{k}$) instead of its functional margin $\vec{a}_{t}\cdot\vec{y}_{k}$. The function $C(t)$ is given by $C(t)=b t^{1-\epsilon}\left\|\vec a_t \right\|^{-1}$ for the $t$-margitron and by $C(t)=b \left\|\vec a_t \right\|^{-\epsilon}$ for the $\ell$-margitron. We expect that $\epsilon<1$ will result in an enhancement of the margin threshold $C(t)$ relative to the case $\epsilon=1$ (Perceptron with margin) and that this enhancement will eventually lead to a slower average fall off of $C(t)$ with $t$ progressing instead of a genuine increase which is desirable in order for the algorithm to converge. This expectation is further supported by the fact that, as we demonstrate below, $C(t)\le ct^{-\epsilon}$ with $c>0$. Hopefully, such a slower decrease of the margin required by the misclassification condition will ensure convergence to solutions possessing margins which are larger fractions of $\gamma_{\rm d}$.

Taking the inner product of (\ref{update}) with the optimal direction $\vec u$ we obtain
\[
\vec a_{t+1}\cdot \vec u -\vec a_{t}\cdot \vec u =\vec y_k \cdot \vec u \ge \gamma_{\rm d}
\]
a repeated application of which gives \cite{Nov} 
\begin{equation}
\label{lbound}
\left\| \vec a_t \right\| \ge \vec a_t \cdot \vec u \ge \gamma_{\rm d}t \enspace.
\end{equation}
Using (\ref{lbound}) we get $C(t)\le ct^{-\epsilon}$ with $c=b\gamma_{\rm d}^{-1}$ and $c=b\gamma_{\rm d}^{-\epsilon}$ for the $t$- and the $\ell$-margitron, respectively.

\section{Theoretical Analysis}
\begin{lemma}
\label{lemma1}
Let
\[ 
g(t)=t^{\epsilon}-\alpha t^{\epsilon-1}- \beta
\]
with $t\in [1, +\infty)$, $\epsilon >0$, $\; \alpha \ge 1$ and $\beta>0$. Then, there is a single value $t_{\rm b}$ of $t$ satisfying
\[
g(t_{\rm b})=0
\]
which is bounded as follows
\[
\begin{array}{c c c c c}
\hspace{0.3cm} & \alpha+\beta^{\vic{\frac{1}{\epsilon}}} \le t_{\rm b} \le \shh{\frac{1}{\epsilon}}\alpha+\beta^{\vic{\frac{1}{\epsilon}}} & \hspace{0.3cm} &  \epsilon \le 1 &  \\
\hspace{0.3cm} & \shh{\frac{1}{\epsilon}}\alpha+\beta^{\vic{\frac{1}{\epsilon}}} < t_{\rm b} < \alpha+\beta^{\vic{\frac{1}{\epsilon}}}  & \hspace{0.3cm} &\epsilon>1 & .
\end{array}
\]
\end{lemma}
\begin{proof}

The function $g(t)$ with $g(1)< 0$ is unbounded from above and is either strictly increasing (if $\epsilon \le 1$) or has at most one local minimum (if $\epsilon >1$). Therefore, there is a single root $t_{{\rm b}}$ of $g(t)$. In addition, for $g(t)\neq 0$ $\mbox{sign}({\vic t-\vic t_{{\rm b}}})=\mbox{sign}({\vic g(t)})$.
Let $0< \epsilon < 1$. We have $g(\alpha+\beta^{\vic{\frac{1}{\epsilon}}})={\beta}^{\vic{\frac{1}{\epsilon}}}(\alpha+\beta^{\vic{\frac{1}{\epsilon}}})^{\epsilon-1}-\beta
< \beta^{\vic{\frac{1}{\epsilon}}}\beta^{\vic{\frac{\epsilon-1}{\epsilon}}}-\beta=0$,
implying that $t_{\rm b} > \alpha+\beta^{\vic{\frac{1}{\epsilon}}}$. Moreover,
$g(\shh{\frac{1}{\epsilon}}\alpha+\beta^{\vic{\frac{1}{\epsilon}}})
=(\shh{\frac{1-\epsilon}{\epsilon}}\alpha+\beta^{\vic{\frac{1}{\epsilon}}})(\shh{\frac{1}{\epsilon}}\alpha+\beta^{\vic{\frac{1}{\epsilon}}})^{\epsilon -1}-\beta=\beta(1+\shh{\frac{1-\epsilon}{\epsilon}}\alpha\beta^{-\vic{\frac{1}{\epsilon}}})(1+\shh{\frac{1}{\epsilon}}\alpha\beta^{-\vic{\frac{1}{\epsilon}}})^{\epsilon -1}-\beta > \beta(1+\shh{\frac{1}{\epsilon}}\alpha\beta^{-\vic{\frac{1}{\epsilon}}})^{1-\epsilon}(1+\shh{\frac{1}{\epsilon}}\alpha\beta^{-\vic{\frac{1}{\epsilon}}})^{\epsilon -1}-\beta=0$ implying that $t_{\rm b} < \shh{\frac{1}{\epsilon}}\alpha+\beta^{\vic{\frac{1}{\epsilon}}}$. (Here we make use of $1+ q z > (1+z)^q$ for $-1<z\neq 0$ and $0<q<1$.) If $\epsilon >1$, instead, both the above inequalities are reversed. (Here we make use of $1-q z < (1+z)^{-q}$ for $z,q>0$.) Finally, for $\epsilon=1$ obviously $t_{\rm b}=\alpha+\beta$.\medskip \hfill \fbox{}
\end{proof}
\begin{theorem}
\label{theorem1}
The t-margitron with $0 < \epsilon \le 1$ converges in  
\begin{eqnarray}
\label{tmartbound}
& t_{\rm c} \le \shh{\frac{1}{\epsilon}}\frac{\vic R^2}{\vic \gamma^2_{\rm d}}+\left(\shh{\frac{2}{2-\epsilon}}\frac{\vic b}{\vic \gamma^2_{\rm d}}\right)^{\vic{\frac{1}{\epsilon}}} 
\end{eqnarray}
updates to a solution hyperplane possessing directional margin $\gamma^\prime_{\rm d}$ which is a fraction $f$ of the maximum directional margin $\gamma_{\rm d}$ obeying the inequality
\begin{eqnarray}
\label{tmarfrbound}
& f \equiv \frac{\vic \gamma^\prime_{\rm d}}{\vic \gamma_{\rm d}}\ge \left(\frac{\vic R^2}{\vic b}+\shh{\frac{ 2}{2-\epsilon}} \right)^{-1} \enspace.
\end{eqnarray}
Moreover, an after-running estimate of $\frac{\vic \gamma^\prime_{\rm d}}{\vic \gamma_{\rm d}}$ is obtainable from 
\begin{equation}
\label{tmarfrest}
\begin{array}{c}
\frac{\vic \gamma^\prime_{\rm d}}{\vic \gamma_{\rm d}}\ge f_{\rm est} \equiv \left(\frac{\vic R^2}{\vic b}t_{\rm c}^{\epsilon-1}+\shh{\frac{2}{2-\epsilon}} \right)^{-1} \enspace.
\end{array}
\end{equation}
\end{theorem}
\begin{proof}

From (\ref{update}) and taking into account (\ref{misclast}) we get
\[
\left\|\vec{a}_{t+1}\right\|^2- \left\|\vec{a}_{t}\right\|^2 =\left \|\vec{y}_{k}\right \|^2
+2\vec{y}_k \cdot \vec{a}_{t} \le R^2+2bt^{1-\epsilon}
\]
a repeated application $t$ times of which leads to
\begin{eqnarray}
\label{tmarupbound}
\left\|\vec{a}_{t}\right\|^2 \le R^2 t+2b\sum^{t-1}_{l=1}l^{1-\epsilon}
\le R^2t+2b \int^t_0 l^{1-\epsilon}dl=R^2t+\shh{\frac{2}{2-\epsilon}}bt^{2-\epsilon}\enspace.
\end{eqnarray}
Combining (\ref{lbound}) with (\ref{tmarupbound}) we obtain
\begin{eqnarray}
\label{tmarsq}
&\gamma_{\rm d}t \le \left\|\vec{a}_{t}\right\| \le R\sqrt{t+\shh{\frac{2}{2-\epsilon}}\frac{\vic b}{\vic R^2}t^{2-\epsilon}} 
\end{eqnarray}
from where
\begin{eqnarray}
\label{tineq1}
& t^{\epsilon} \le \frac{\vic R^2}{\vic \gamma^2_{\rm d}}t^{\epsilon-1}+\shh{\frac{2}{2-\epsilon}}\frac{\vic b}{\vic \gamma^2_{\rm d}}
\end{eqnarray}
or, equivalently, 
\begin{eqnarray}
\label{tineq}
& g(t) \equiv t^{\epsilon}-\frac{\vic R^2}{\vic \gamma^2_{\rm d}}t^{\epsilon-1}-\shh{\frac{2}{2-\epsilon}}\frac{\vic b}{\vic \gamma^2_{\rm d}} \le 0 \enspace. 
\end{eqnarray}
The value $t_{\rm b}$ of t for which the above relation holds as an equality provides an upper bound on the number of updates $t_{\rm c}$ required for convergence. According to Lemma \ref{lemma1} there is a single such value which is bounded as stated there. This leads to the looser bound of (\ref{tmartbound}).  

Combining (\ref{misclast}) with (\ref{nmisclas}) and using (\ref{tmarupbound}) we obtain
\begin{eqnarray}
\label{tfrac1}
& \frac{\vic{C(t)}}{\vic \gamma_{\rm d}}=\frac{\vic {b \: t^{1-\epsilon}}}{\vic \gamma_{\rm d}\left\|\vec a_t \right\|} \ge \left( \frac{\vic {\gamma_{\rm d}R}}{\vic b}\sqrt{t^{2\epsilon-1}+\shh{\frac{2}{2-\epsilon}}\frac{\vic b}{\vic R^2}t^{\epsilon}} \right)^{-1} \enspace .
\end{eqnarray}
Multiplying both sides of (\ref{tineq1}) with its r.h.s. we get
\begin{eqnarray}
& t^{\epsilon}\left(\frac{\vic R^2}{\vic \gamma^2_{\rm d}}t^{\epsilon-1}+\shh{\frac{2}{2-\epsilon}}
\frac{\vic b}{\vic \gamma^2_{\rm d}}\right) \le \left(\frac{\vic R^2}{\vic \gamma^2_{\rm d}}t^{\epsilon-1}+\shh{\frac{2}{2-\epsilon}}\frac{\vic b}{\vic \gamma^2_{\rm d}}\right)^2 \enspace, \nonumber
\end{eqnarray}
or
\begin{eqnarray}
& \frac{\vic {\gamma_{\rm d}R}}{\vic b}\sqrt{t^{2\epsilon-1}+\shh{\frac{2}{2-\epsilon}}\frac{\vic b}{\vic R^2}t^{\epsilon}} \le 
\frac{\vic R^2}{\vic b}t^{\epsilon-1}+\shh{\frac{2}{2-\epsilon}}\enspace. \nonumber
\end{eqnarray}
Using this last inequality and taking into account that $f={\vic \gamma^\prime_{\rm d}}/{\vic \gamma_{\rm d}}\ge {\vic{C(t_{\rm c})}}/{\vic \gamma_{\rm d}}$ (\ref{tfrac1}) leads to (\ref{tmarfrest}).
Setting $t_{\rm c}=1$ in (\ref{tmarfrest}) we obtain the weaker bound of (\ref{tmarfrbound}). \medskip \hfill \fbox{}
\end{proof}

\begin{remark}
Noticing that the number of updates $t_{\rm c}$ required for convergence of the t-margitron satisfies (\ref{tineq1}) we get
\begin{eqnarray}
&\gamma_{\rm d} \le R\sqrt{t^{-1}_{\rm c}+\shh{\frac{2}{2-\epsilon}}\frac{\vic b}{\vic R^2}t^{-\epsilon}_{\rm c}} \nonumber
\end{eqnarray}
from where an alternative after-running lower bound on ${\vic \gamma^\prime_{\rm d}}/{\vic \gamma_{\rm d}}$ is obtainable. This bound, however, does not have to be smaller than $1-\frac{\epsilon}{2}$.
\end{remark}

\begin{remark}
The r.h.s. of (\ref{tfrac1}) has in the interval $[1, +\infty)$ a single extremum, which is a maximum, at $\vic t_{\star}=\left({|1-2\epsilon|(2-\epsilon)}{(2\epsilon)^{-1}}\frac{\vic R^2}{\vic b}\right)^{\vic{\frac{1}{1-\epsilon}}}\mbox{sign}(1-2\epsilon)$.
Therefore, it is legitimate in calculating a lower bound on ${\vic{C(t_{\rm c})}}/{\vic \gamma_{\rm d}}$ using (\ref{tfrac1}) to replace $t_{\rm c}$ with $t_{\rm b}$ provided $t_{\rm c} \ge t_{\star}$. This leads to the stronger than the one of (\ref{tmarfrbound}) bound
\begin{eqnarray}
\label{tboundstr}
& f\ge \left(\frac{\vic R^2}{\vic b}t^{\epsilon-1}_{\rm b}+\shh{\frac{2}{2-\epsilon}} \right)^{-1} 
\end{eqnarray}
which, however, is $\gamma_{\rm d}$-dependent. The condition $t_{\rm c} \ge t_{\star}$ is automatically satisfied for $\frac{1}{2} \le \epsilon \le 1$. For $0 < \epsilon < \frac{1}{2}$, instead, we may ensure that $t_{\rm c} \ge t_{\star}$ if the r.h.s. of (\ref{tfrac1}) is larger than or equal to 1 for $t=1$ and as a consequence the normalised margin threshold $C(t)$ is initially not lower than the maximum directional margin $\gamma_{\rm d}$, i.e. $C(1) \ge \gamma_{\rm d}$. A condition sufficient for this to be the case is $\frac{\vic b}{\vic R^2} \ge \frac{\vic \gamma_{\rm d}}{\vic R}\left(1+\shh{\frac{2}{2-\epsilon}} \frac{\vic \gamma_{\rm d}}{\vic R} \right)$. In this event the algorithm is forced to converge only after $C(t)$ has fallen bellow $\gamma_{\rm d}$ which cannot occur as long as $t < t_{\star}$.
If we choose
\begin{equation}
\label{bt}
\frac{\vic b}{\vic R^2}=\left(1-\frac{\epsilon}{2}\right)^{1-\epsilon}\vic\delta^{-\epsilon}\left(\frac{\vic \gamma^2_{\rm d}}{\vic R^2}\right)^{1-\epsilon}
\end{equation}
and replace in (\ref{tboundstr}) $t_{\rm b}$ with its lower bound $t_{\rm lb}\equiv \left(\shh{\frac{2}{2-\epsilon}}\frac{\vic b}{\vic \gamma^2_{\rm d}}\right)^{\vic{\frac{1}{\epsilon}}}
=\shh{\frac{2}{2-\epsilon}}\delta^{-1}\frac{\vic R^2}{\vic \gamma^2_{\rm d}}$, which is lower than the lower bound inferred from Lemma \ref{lemma1}, we can easily verify that $f\ge \left(\delta+\shh{\frac{2}{2-\epsilon}} \right)^{-1}$. If $0 < \epsilon < \frac{1}{2}$ the parameter $\delta$ should satisfy the constraint $\delta \le \left(1-\shh{\frac{\epsilon}{2}}\right)^{\vic{\frac{1-\epsilon}{\epsilon}}}\left(\frac{\vic \gamma_{\rm d}}{\vic R}\right)^{\vic{\frac{1}{\epsilon}}-2}\left(1+\shh{\frac{2}{2-\epsilon}} \frac{\vic \gamma_{\rm d}}{\vic R} \right)^{-\vic{\frac{1}{\epsilon}}}$ which for $0 < \epsilon \ll \frac{1}{2}$ and $\frac{\vic \gamma_{\rm d}}{\vic R}\ll 1$ suggests a rather slow convergence. Thus, it is not advisable in this case to employ values of $b$ for which the constraint on $\delta$ is satisfied. The algorithm will still be able to achieve a large fraction of $\gamma_{\rm d}$ if it happens to converge in a sufficiently large number of updates $t_{\rm c}$ as it can be deduced from (\ref{tmarfrest}).


\end{remark}

\begin{lemma}
\label{lemma2}
For $x, y > 0$ and $-1 < \epsilon \le 1$ it holds that
\begin{eqnarray}
\label{ineq}
& \frac{\vic{x^{1+\epsilon}}}{\vic{1+\epsilon}}-\frac{\vic {y^{1+\epsilon}}}{\vic{1+\epsilon}}\le \frac{\vic{ x^2- y^2}}{\vic{2 y^{1-\epsilon}}}\enspace.
\end{eqnarray}
\end{lemma}
\begin{proof} 

For $\epsilon=1$ (\ref{ineq}) holds obviously as an equality. For $-1 < \epsilon < 1$ (\ref{ineq}) is equivalent to $1 \le \shh{\frac{1+\epsilon}{2}}\alpha^{1-\epsilon}+\shh{\frac{1-\epsilon}{2}}\alpha^{-(1+\epsilon)}$,
with $\alpha=x/y$. The r.h.s. of the above inequality is minimised for $\alpha=1$ and takes the value 1. 
\medskip \hfill \fbox{}
\end{proof}
\begin{lemma}
\label{lemma3}
For $t \ge 1$ and $0<\epsilon \le 1$ it holds that
\begin{eqnarray}
\label{log}
& \frac{\vic{t^{\epsilon}-1}}{\vic{\epsilon}} \le t^{\epsilon}(\ln t)^{1-\epsilon}-[\epsilon] \enspace,
\end{eqnarray}
where $[\epsilon]$ denotes the integer part of $\epsilon$.
\end{lemma}
\begin{proof}

For $t=1$ or $\epsilon=1$ (\ref{log}) holds obviously as an equality. Let $t>1$ and $0<\epsilon<1$. Then, with $x=t^\epsilon$ (\ref{log}), as a strict inequality, is equivalent to $f(\epsilon)=\epsilon^{\epsilon}x(\ln x)^{1-\epsilon}-x+1>0$.
For $x \ge e^e$ we have $\frac{\vic {df}}{\vic {d\epsilon}}<0$ from where $f(\epsilon)> \lim \limits_{\epsilon \to 1}f(\epsilon)=1$.
For $1<x< e^e$, instead, $f$ has only one local minimum at $\epsilon=e^{-1}\ln x$ with value at that minimum given by
$h(x)=x^{\shhh{\frac{\vic {e-1}}{\vic e}}}\ln x -x +1$. It can be easily shown that $\frac{\vic {dh}}{\vic {dx}}=(1-e)x^{-e^{-1}}\ln x^{-e^{-1}}+x^{-e^{-1}}-1 $ has no local minima in the interval $(1,e^e)$. Thus, 
$\frac{\vic {dh}}{\vic {dx}}>\min\left\{\lim\limits_{x \to 1} \frac{\vic {dh}}{\vic {dx}}, 
\lim\limits_{x \to e^e}\frac{\vic {dh}}{\vic {dx}} \right\}=0$. Therefore, $h(x)> \lim\limits_{x \to 1}h(x)=0$ and consequently $f(\epsilon)>0$. \medskip \hfill \fbox{}
\end{proof}
\begin{lemma}
\label{lemma4}
Let 
\begin{eqnarray}
& g(t)=t^{\epsilon}-\Bigl(\alpha_1\left(\frac{\vic \ln \vic t}{\vic t} \right)^{1-\epsilon}+\alpha_2 t^{-1}\Bigr)-\beta
\nonumber
\end{eqnarray}
with $t\in[1, +\infty)$, $0< \epsilon < 1$,$\;\; \alpha_1, \alpha_2, \beta>0$  and $\;\; \alpha \equiv \alpha_1+\alpha_2 \ge 2+\epsilon$. Then, $g(t_0)>0$ with
\begin{eqnarray}
t_0 = (\shh{\frac{1}{\epsilon}}\alpha+\beta^{\vic{\frac{1}{\epsilon}}})
\biggl(\ln(\shh{\frac{1}{\epsilon}}\alpha+\beta^{\vic{\frac{1}{\epsilon}}})\biggr)^{1-\epsilon} \enspace. \nonumber
\end{eqnarray}
\end{lemma}
\begin{proof}

Let $\lambda ={{\alpha}/(\alpha +\epsilon\beta^{\vic{\frac{1}{\epsilon}}}})<1$ and $x=\ln\srr{\frac{\alpha}{\lambda \epsilon}}\ge \zeta \equiv \ln \left(1+\shh{\frac{2}{\epsilon}}\right)> 1$ such that  $t_0=\srr{\frac{\alpha}{\lambda \epsilon}}x^{1-\epsilon}\ge 1+\shh{\frac{2}{\epsilon}}>e $, $\ln t_0=x+(1-\epsilon)\ln x >1$, $\beta={\alpha}^{\epsilon}(1-\lambda)^{\epsilon}/(\lambda\epsilon)^{\epsilon}$ and $\alpha_2 t_0^{-1} < \alpha_2 \left({\vic{\ln t_0}}/{\vic{t_0}}\right)^{1-\epsilon}$. Then,
\[
\begin{array}{c}
g(t_0)> t_0^{\epsilon}-\alpha\left(\frac{\vic{\ln t_0}}{\vic{t_0}}\right)^{1-\epsilon}-\beta
=t^{\epsilon}_0{\biggl(1-\lambda{\epsilon}\left({1+(1-\epsilon)}\frac{\vic{\ln x}}{\vic x}\right)^{1-\epsilon}
-\frac{\vic {\left(1- \lambda\right)^{\epsilon}}}{\vic{x^{(1-\epsilon)\epsilon}}} \biggr)}
\end{array}
\]
\[
\begin{array}{@{}l}
> 1-\lambda\epsilon(1+(1-\epsilon)e^{-1})^{1-\epsilon}-\left(1-\lambda \right)^{\epsilon}\zeta^{(\epsilon-1)\epsilon}
\enspace .
\end{array}
\]
Here we made use of $t^{\epsilon}_0>1$, $\voc{\frac{\ln x}{x}}\le e^{-1}$ and $\frac{\vic 1}{\vic{x^{(1-\epsilon)\epsilon}}} \le \zeta^{(\epsilon-1)\epsilon}$. This last expression is minimised with respect to $\lambda$ for $\lambda=\frac {\vic{1+(1-\epsilon)e^{-1}-\zeta^{-\epsilon}}}{\vic{1+(1-\epsilon)e^{-1}}}$ which substituted leads to 
\[
g(t_0) > (1+(1-\epsilon)e^{-1})^{-\epsilon}f(\epsilon) 
\]
with $f(\epsilon) \equiv(1-\epsilon)(1-\zeta^{-\epsilon})+(1+(1-\epsilon)e^{-1})^{\epsilon}-(1+\epsilon(1-\epsilon)e^{-1})$. 
Employing the expansion $\ln z=\sum^{\infty}_{k=1}\frac{2}{2k-1}\left(\frac{z-1}{z+1}\right)^{2k-1}$ for $z>0$ \cite{Grad}
we obtain $\zeta >\left(\shh{ \frac{1+\epsilon}{2}}\right)^{-1}$ from where $\zeta^{-\epsilon} <\left(\shh{ \frac{1+\epsilon}{2}}\right)^{\epsilon}=\left(1-\shh{\frac{1}{2}}(1-\epsilon)\right)^{\epsilon}<1-\shh{\frac{1}{2}}\epsilon(1-\epsilon)$. Moreover, $(1+(1-\epsilon)e^{-1})^{\epsilon}-(1+\epsilon(1-\epsilon)e^{-1})>-\shh{\frac{1}{2}}\epsilon(1-\epsilon)^{3}e^{-2}$
since $(1+z)^q-(1+qz)>\shh{\frac{1}{2}}q(q-1)z^2$ for $z>0$ and $0<q<1$. Thus, $f(\epsilon)>\shh{\frac{1}{2}}\epsilon(1-\epsilon)^{2}(1-(1-\epsilon)e^{-2})>0$ leading to $g(t_0)>0$.
\medskip \hfill \fbox{}

\end{proof}

\begin{theorem}
\label{theorem2}
The $\ell$-margitron with $0 < \epsilon \le 1$ converges in
\begin{eqnarray}
\label{lmartbound}
t_{\rm c} \le (\shh{\frac{1}{\epsilon}}A+B^{\vic{\frac{1}{\epsilon}}})
\left(\ln(\shh{\frac{1}{\epsilon}}A+B^{\vic{\frac{1}{\epsilon}}})\right)^{1-\epsilon}
\end{eqnarray}
updates, with $A=(2+\epsilon - 2[\epsilon])\frac{\vic R^2}{\vic \gamma^2_{\rm d}}$ and $B=(1+\epsilon)\frac{\vic b}{\vic \gamma^{\shri{1+\epsilon}}_{\rm d}}$, to a solution hyperplane possessing directional margin $\gamma^\prime_{\rm d}$ which is a fraction $f$ of the maximum directional margin $\gamma_{\rm d}$ obeying the inequality
\begin{equation}
\label{lmarfrbound}
\begin{array}{@{}l}
f \equiv \frac{\vic \gamma^\prime_{\rm d}}{\vic \gamma_{\rm d}} \ge {\biggl\{ \inrr{\frac{(1+\epsilon)^{2\epsilon-[\epsilon]}}{(2\epsilon)^{\epsilon}}}\left(\frac{\vic R^{\shri{1+\epsilon}}}{\vic b}\right)+ 1+\epsilon \biggr\}^{-1} } \enspace.
\end{array}
\end{equation}
Moreover, for $0<\epsilon<1$ an after-running estimate of $\frac{\vic \gamma^\prime_{\rm d}}{\vic \gamma_{\rm d}}$ is obtainable from 
\begin{equation}
\label{lmarfrest}
\begin{array}{@{}l}
\frac{\vic \gamma^\prime_{\rm d}}{\vic \gamma_{\rm d}} \ge f_{\rm est}\equiv \biggl\{\frac{\vic R^{\shri{1+\epsilon}}}{\vic b}\Bigl(N^{1+\epsilon}+ \shh{\frac{1+\epsilon}{2\epsilon}}
\left(\frac{\vic R}{\vic \gamma^{\prime}_{\rm d}}\right)^{1-\epsilon}\bigl(\vic t_{\rm c}^{\epsilon} - \frac{\vic {N-\epsilon}}{\vic N^{\shri{1-\epsilon}}}\bigr)\Bigr)t^{-1}_c+ 1+\epsilon  \biggr\}^{-1} \enspace . \\
\end{array}
\end{equation}
Here the integer $N>0$ satisfies any of the constraints 
\begin{eqnarray}
\label{lmarfrest3}
&t_{\rm c}\ge N \ge \shh{\frac{1+\epsilon}{2}}\left(\frac{\vic R}{\vic \gamma^\prime_{\rm d}}\right)^{1-\epsilon}, \:\: t_{\rm c} \ge N \left(\frac{\vic{1-\epsilon N^{-1}}}{\vic{1-\epsilon}} \right)^{\vic{\frac{1}{\epsilon}}}.
\end{eqnarray}
Obviously, the choice $N=1$ is always acceptable. A near optimal choice of $N$ is $N_{\rm opt}=\voc{\left[\shh{\frac{1}{2}}\left(\inc{\frac{\vic R}{\vic \gamma^\prime_{\rm d}}}\right)^{1-\epsilon}\right]}+1$, provided it satisfies one of the above constraints.
\end{theorem}
\begin{proof}

From (\ref{update}) and taking into account (\ref{misclasl}) we get
\[
\left\|\vec{a}_{t+1}\right\|^2- \left\|\vec{a}_{t}\right\|^2 =\left \|\vec{y}_{k}\right \|^2
+2\vec{y}_k \cdot \vec{a}_{t}  \le R^2+2 b\left\|\vec a_t \right\|^{1-\epsilon}
\]
or, assuming $t \ge 1$,
\begin{eqnarray}
& \frac{\left\|\vec{a}_{t+1}\right\|^2-\left\|\vec{a}_{t}\right\|^2}{2\left\|\vec{a}_{t}\right\|^{\shri{1-\epsilon}}}
\le \frac{\vic R^2}{2\left\|\vec{a}_{t}\right\|^{\shri{1-\epsilon}}}+\vic b \enspace . \nonumber
\end{eqnarray}
By using (\ref{lbound}) in the r.h.s. of the above inequality and (\ref{ineq}) in its l.h.s. we obtain
\[
\begin{array}{@{}l}
\frac{\left\|\vec{a}_{t+1}\right\|^{\shri{1+\epsilon}}}{1+\epsilon}-\frac{\left\|\vec{a}_{t}
\right\|^{\shri{1+\epsilon}}}{1+\epsilon} \le \frac{1}{2}\frac{\vic R^2}{\vic \gamma_{\rm d}^{\shri{1-\epsilon}}} t^{\epsilon-1}+b 
\end{array}
\]
a repeated application $t-N$ times ($t > N \ge 1$) of which gives
\[
\begin{array}{@{}l }
\frac{\left\|\vec{a}_{t}\right\|^{\shri{1+\epsilon}}}{1+\epsilon}-\frac{\left\|\vec{a}_N
\right\|^{\shri{1+\epsilon}}}{1+\epsilon}\le \frac{1}{2}\frac{\vic R^2}{\vic \gamma^{\shri{1-\epsilon}}_{\rm d}}\sum \limits^{t-1}_{l=N}l^{\epsilon-1}
+b(t-N) \\
\end{array}
\]
\[
\begin{array}{c c}
\hspace{0.3cm} & \le \frac{1}{2}\frac{\vic R^2}{\vic \gamma^{\shri{1-\epsilon}}_{\rm d}}\left(N^{\epsilon-1}+\int \limits^{t-1}_{l=N}l^{\epsilon-1}dl\right)
+bt \\
\end{array}
\]
\[
\begin{array}{c c}
\hspace{1cm} & = \frac{1}{2}\frac{\vic R^2}{\vic \gamma^{\shri{1-\epsilon}}_{\rm d}}\left(N^{\epsilon-1}+\frac{1}{\epsilon}(t-1)^{\epsilon} -\frac{1}{\epsilon}N^\epsilon\right)+bt \\
\end{array}
\]
\[
\begin{array}{c c}
\hspace{1cm} & \le \frac{1}{2}\frac{\vic R^2}{\vic \gamma^{\shri{1-\epsilon}}_{\rm d}}
\left(N^{\epsilon-1}+\frac{1}{\epsilon}t^{\epsilon} -[\epsilon]-\frac{1}{\epsilon}N^\epsilon\right)+bt \enspace . 
\end{array}
\]
Thus, employing the obvious bound $\left\|\vec{a}_{N}\right\|\le RN$, we are led to
\begin{equation}
\label{lmarupbound}
\left\|\vec{a}_{t}\right\| \le {L_N}_t \equiv R\biggl( N^{1+\epsilon}
+\shh{\frac{1+\epsilon}{2\epsilon}}\left(\frac{\vic R}{\vic \gamma_{\rm d}}\right)^{1-\epsilon}\Bigl(t^{\epsilon} -\frac{\vic{N- \epsilon+[\epsilon]}}{\vic N^{1-\epsilon}}\Bigr)+(1+\epsilon)\frac{\vic b}{\vic R^{1+\epsilon}}t \biggr)^{\vic{\frac{1}{1+\epsilon}}} 
\end{equation}
which, although derived for $t>N$, turns out to be satisfied even for $t=N$.
Combining (\ref{lbound}) with (\ref{lmarupbound}) we obtain
\begin{eqnarray}
\label{sq1}
&\gamma_{\rm d}^{1+\epsilon}t^{1+\epsilon} \le \left\|\vec{a}_{t}\right\|^{1+\epsilon} \le ({L_N}_t)^{1+\epsilon}
\end{eqnarray}
from where
\begin{eqnarray}
\label{sq2}
& t^{\epsilon} \le \left(\frac{\vic {L_N}_t}{\vic \gamma_{\rm d}}\right)^{1+\epsilon}t^{-1}
\end{eqnarray}
or, equivalently,
\begin{equation}
\label{sq3}
\vic g_N(t) \le 0
\end{equation}
with
\begin{equation}
\label{gn}
\begin{array}{l}
\vic g_N(t)\equiv t^{\epsilon} - \left(\frac{\vic {L_N}_t}{\vic \gamma_{\rm d}}\right)^{1+\epsilon}t^{-1} \\\hspace{-1pt} =\hspace{-1pt} t^{\epsilon}\hspace{-1pt}-\hspace{-1pt}\left(\frac{\vic R}{\vic \gamma_{\rm d}}\right)^{1+\epsilon}N^{1+\epsilon}\vic t^{-1}
\hspace{-1pt}-\hspace{-1pt}\shh{\frac{1+\epsilon}{2\epsilon}}\frac{\vic R^2}{\vic \gamma^2_{\rm d}}\left(t^{\epsilon} \hspace{-1pt}-\hspace{-1pt} N^\epsilon+ (\epsilon-[\epsilon])N^{\epsilon-1}\right)t^{-1}\hspace{-1pt}-\hspace{-1pt}(1+\epsilon)\frac{\vic b}{\vic \gamma^{\shri{1+\epsilon}}_{\rm d}}\enspace.
\end{array}
\end{equation}
Let us consider the derivative of $\vic g_N(t)$
\begin{eqnarray}
& \frac{\vic d \vic g_N}{\vic {dt}}=  D_N(t)\: t^{-2} \enspace ,\nonumber
\end{eqnarray}
where
\[
\begin{array}{c}
D_N(t)= \epsilon\: t^{1+\epsilon}+ \left(\frac{\vic R}{\vic \gamma_{\rm d}}\right)^{1+\epsilon}
N^{1+\epsilon}+\shh{\frac{1+\epsilon}{2\epsilon}}\frac{\vic R^2}{\vic \gamma^2_{\rm d}}\Bigl((1-\epsilon)t^{\epsilon} - N^\epsilon+ (\epsilon-[\epsilon])N^{\epsilon-1}\Bigr).
\end{array}
\]
$D_N(t)$ is strictly increasing and therefore has at most one root $t_{{\rm r}_N}$ ($D_N(t_{{\rm r}_N})=0$) where obviously $\vic g_N(t)$ acquires a minimum (since $\vic g_N(t)$ is unbounded from above) with $\vic g_N(t_{{\rm r}_N})<0$ (since $\vic g_N(N)<0$).
Thus, $\vic g_N(t)$ starts from negative values at $t=N$ and with $t$ increasing either tends monotonically to infinity or decreases further until it acquires a minimum at $t=t_{{\rm r}_N}$ and then increases monotonically towards infinity. In both cases there is a single value $t_{{\rm b}_N}$ of $t$ for which
\begin{equation}
\label{ltb}
\vic g_N(t_{{\rm b}_N})=0
\end{equation}
and moreover for $\vic g_N(t) \neq 0$
\begin{eqnarray}
\label{sign}
& \mbox{sign}({\vic t-\vic t_{{\rm b}_N}})=\mbox{sign}({\vic g_N(\vic t)}) \enspace.
\end{eqnarray}
The unique value $t_{{\rm b}_N}$ of t for which (\ref{sq1}), (\ref{sq2}) and (\ref{sq3}) hold as equalities provides an upper bound on the number of updates $t_{\rm c}$ required for convergence.

Combining (\ref{misclasl}), (\ref{nmisclas}), (\ref{lmarupbound}), (\ref{gn}) and (\ref{ltb}) we get
\begin{equation}
\label{lfractb}
\begin{array}{@{}l}
f=\frac{\vic \gamma^\prime_{\rm d}}{\vic \gamma_{\rm d}}\ge \frac{\vic{C(t_{\rm c})}}{\vic \gamma_{\rm d}} 
= \frac{\vic b}{\vic \gamma_{\rm d}\vic{\left\|\vec a_{t_{\rm c}} \right\|^{\epsilon}}}\ge \frac{\vic b}{\vic \gamma_{\rm d} \vic {L^{\epsilon}_N}_{t_{\rm c}} } \ge \frac{\vic b}{\vic \gamma_{\rm d} \vic {L^{\epsilon}_N}_{t_{{\rm b}_N}} }=
\frac{\vic b}{\vic \gamma_{\rm d}^{\shri{1+\epsilon}}\vic{t^{\epsilon}_{{\rm b}_N}}} \vspace{0.1cm} \\ 
=\biggl\{\frac{\vic R^{\shri{1+\epsilon}}}{\vic b }\biggl(N^{1+\epsilon}+ \shh{\frac{1+\epsilon}{2\epsilon}}
\left(\frac{\vic R}{\vic \gamma_{\rm d}}\right)^{1-\epsilon}\Bigl(t^{\epsilon}_{{\rm b}_N} \hspace{-2pt}-\frac{\vic{N- \epsilon+[\epsilon]}}{\vic N^{\shri{1-\epsilon}}}\Bigr)\biggr)t^{-1}_{{\rm b}_N}+ 1+\epsilon \biggr\}^{-1} \enspace .\\
\end{array}
\end{equation}
For $\epsilon=1$ the above lower bound on $f$ is optimised for $N=1$ in which case it reduces to (\ref{lmarfrbound}). For $0<\epsilon<1$ we may replace in the above lower bound on $f$ first $\vic \gamma_{\rm d}$ with $\vic \gamma^{\prime}_{\rm d}$ and subsequently, on the condition that one of the constraints (\ref{lmarfrest3}) is satisfied, ${t_{{\rm b}_N}}$ with $t_{\rm c}$ since both replacements can be shown to loosen the bound. Thus, we obtain $f \ge f_{\rm est}$ with $f_{\rm est}$ given by (\ref{lmarfrest}). An approximate maximisation of $f_{\rm est}$ with respect to $N$ leads to the near optimal value $N_{\rm opt}$ of Theorem \ref{theorem2}.

Let us choose $N=1$ in (\ref{gn}) and replace ${\left(\shh{\frac{\vic R}{\vic \gamma_{\rm d}}}\right)^{1+\epsilon}}$ with ${\frac{\vic R^2}{\vic \gamma^2_{\rm d}}}$, thereby lowering the value of $\vic g_1(t)$
\begin{equation}
\label{appr1}
\begin{array}{@{}l}
\vic g_1(t)\ge t^{\epsilon}
-\shh{\frac{1+\epsilon}{2}}\frac{\vic R^2}{\vic \gamma^2_{\rm d}}\Bigl(\frac{\vic{t^{\epsilon}-1}}{\vic{\epsilon}} +\shh{\frac{2}{1+\epsilon}}+ 1-[\epsilon]\Bigr)\frac{1}{\vic t}-(1+\epsilon)\frac{\vic {b}}{\vic \gamma^{\shri{1+\epsilon}}_{\rm d}} .
\end{array}
\end{equation}
By employing (\ref{log}) in the r.h.s. of (\ref{appr1}) we obtain
\begin{equation}
\label{appr2}
\begin{array}{@{}l}
\vic g_1(t)\ge \bar{g}(t) \equiv t^{\epsilon}-\frac{\vic R^2}{\vic \gamma^2_{\rm d}}\Bigl(\shh{\frac{1+\epsilon}{2}}t^{\epsilon}(\ln t)^{1-\epsilon}+ \shh{\frac{3+\epsilon}{2}}(1-[\epsilon])\Bigr)\frac{1}{\vic t}-(1+\epsilon)\frac{\vic {b}}{\vic \gamma^{\shri{1+\epsilon}}_{\rm d}}.
\end{array}
\end{equation}
For $0<\epsilon<1$ $\bar{g}(t)$ becomes a function of the type considered in Lemma \ref{lemma4} with $\alpha=(2+\epsilon)\shh{\frac{\vic R^2}{\vic \gamma^2_{\rm d}}} \ge 2+\epsilon$. Obviously $t_0$ of Lemma \ref{lemma4} satisfies $\vic g_1(t_0)\ge \vic {\bar{g}}(t_0)>0$ and according to (\ref{sign}) is an upper bound on $\vic {t_{{\rm b}_1}}$. Also for $\epsilon=1$ $\bar{g}(t)$ becomes a function of the type considered in Lemma \ref{lemma1} and $t_{\rm b}$ of Lemma \ref{lemma1} is an upper bound on $\vic {t_{{\rm b}_1}}$. Actually in this very special case $\vic {t_{{\rm b}_1}}$ coincides with $t_{\rm b}$ (since (\ref{appr2}) holds as an equality) which, in turn, coincides with its upper and lower bound. This, given that $t_{\rm c} \le\vic {t_{{\rm b}_1}}$, completes the proof of (\ref{lmartbound}).

Alternatively using $-\vic{\frac{1}{\epsilon}} +\frac{2}{1+\epsilon} \le 0$, $\vic{1-[\epsilon]}\le \vic{(1-[\epsilon])}t^\epsilon$ and $(1+\epsilon)(1+\epsilon-[\epsilon])=(1+\epsilon)^{2-[\epsilon]}$ in the r.h.s. of (\ref{appr1}) we obtain
\[
\begin{array}{@{}l}
\vic g_1(t)\ge \tilde{g}(t) \equiv t^{\epsilon}
-\shh{\frac{1}{2\epsilon}}(1+\epsilon)^{2-[\epsilon]}\frac{\vic R^2}{\vic \gamma^2_{\rm d}}\vic t^{\epsilon-1}-
(1+\epsilon)\frac{\vic b}{\vic \gamma^{\shri{1+\epsilon}}_{\rm d}} \enspace .
\end{array}
\]
The function $\tilde{g}(t)$ is of the type considered in Lemma \ref{lemma1} and its only root $\tilde{t}_{\rm b}$ satisfying
\begin{equation}
\label{tbtilde}
\tilde{g}(\tilde{t}_{\rm b})=0 
\end{equation}
is an upper bound on the number of updates looser than $\vic {t_{{\rm b}_1}}$ i.e. $\vic {t_{{\rm b}_1}}\le\tilde{t}_{\rm b}$. Moreover, the upper bound on $\tilde{t}_{\rm b}$ from Lemma \ref{lemma1} is an alternative upper bound on $t_{\rm c}$. Combining (\ref{lfractb}) for $N=1$, the inequality $\vic {t_{{\rm b}_1}}\le\tilde{t}_{\rm b}$ and (\ref{tbtilde}) we obtain

\begin{equation}
\label{appr3}
\begin{array}{c}
{f \ge \frac{\vic b}{\vic \gamma_{\rm d}^{\shri{1+\epsilon}}\vic{t^{\epsilon}_{{\rm b}_1}}}\ge \frac{\vic b}{\vic \gamma_{\rm d}^{\shri{1+\epsilon}}\vic{\tilde{t}^{\epsilon}_{\rm b}}} }
={\biggl\{\frac{\vic R^{\shri{1+\epsilon}}}{\vic b} \shh{\frac{1}{2\epsilon}}(1+\epsilon)^{2-[\epsilon]}
\left(\frac{\vic R}{\vic \gamma_{\rm d}}\right)^{1-\epsilon}\tilde{t}_{\rm b}^{\epsilon-1}+ 1+\epsilon \biggr\}^{-1}}
\enspace .
\end{array}
\end{equation}
Additionally, $\tilde{t}_{\rm lb1} \equiv \shh{\frac{1}{2\epsilon}}(1+\epsilon)^{2-[\epsilon]}\frac{\vic R}{\vic \gamma_{\rm d}}$ is a lower bound on $\tilde{t}_{\rm b}$ since it is lower than the lower bound inferred from Lemma \ref{lemma1}. Replacing $\tilde{t}_{\rm b}$ with its lower bound $\tilde{t}_{\rm lb1}$ in the r.h.s. of (\ref{appr3}) we get the weaker bound (\ref{lmarfrbound}). \medskip \hfill \fbox{}
\end{proof}

\begin{remark}
The lower bounds (\ref{lfractb}) and (\ref{appr3}) on the fraction $f$ involving the unknown maximum margin $\gamma_{\rm d}$ are of great theoretical importance because they guarantee before running that the algorithm will achieve a margin which is a more substantial fraction of $\gamma_{\rm d}$ than the one inferred from (\ref{lmarfrbound}). As a consequence, values of the parameter $b$ smaller than the ones inferred from (\ref{lmarfrbound}) suffice in order for the before-running lower bound on the fraction $f$ to be close to its asymptotic value $(1+\epsilon)^{-1}$. This is quantified in the following theorem.
\end{remark}
\begin{theorem}
\label{theorem3}
The $\ell$-margitron with $0<\epsilon\le 1$ and $b$ (at least as large as the one) given by \vspace{-7pt}
\begin{eqnarray}
\label{bl}
&\frac{\vic b}{\vic R^{\shri{1+\epsilon}}}=\incc{\frac{(1+\epsilon)^{3\epsilon-1-[\epsilon]}}{(2\epsilon\delta)^{\epsilon}}}
\left(\frac{\vic \gamma_{\rm d}}{\vic R}\right)^{1-\epsilon}
\end{eqnarray}
($\delta >0$) converges in a finite number of updates to a solution hyperplane possessing directional margin $\gamma^\prime_{\rm d}$ which is a fraction $f$ of the maximum directional margin $\gamma_{\rm d}$ obeying the inequality
\begin{eqnarray}
\label{blf}
&f=\frac{\vic \gamma^\prime_{\rm d}}{\vic \gamma_{\rm d}}\ge(\delta+1+\epsilon)^{-1} \enspace.
\end{eqnarray}
\end{theorem}
\begin{proof}

Notice that
\begin{eqnarray}
& \tilde{t}_{\rm lb2}\equiv \left((1+\epsilon)\frac{\vic b}{\vic \gamma^{\shri{1+\epsilon}}_{\rm d}}\right)^{\vic {\frac{1}{\epsilon}}}
=\incrr{\frac{(1+\epsilon)^{3-[\epsilon]}}{2\epsilon\delta}}\frac{\vic R^2}{\vic \gamma^2_{\rm d}}
\nonumber
\end{eqnarray}
is a lower bound on $\tilde{t}_{\rm b}$ of (\ref{tbtilde}) since it is lower than the lower bound inferred from Lemma \ref{lemma1}. Replacing $\tilde{t}_{\rm b}$ with its lower bound $\tilde{t}_{\rm lb2}$ in the r.h.s. of (\ref{appr3}) completes the proof. (Larger $b$'s may be regarded as corresponding to smaller $\delta$'s.) \medskip \hfill \fbox{}
\end{proof}

\begin{remark}
\label{remark4}
For $\epsilon \ll 1$ a more accurate determination of $b$ ensuring that (\ref{blf}) holds is obtained from $\frac{\vic b}{\vic R^{1+\epsilon}}$$=\omega^{\epsilon}\left(\frac{\vic \gamma_{\rm d}}{\vic R}\right)^{1-\epsilon}$ with $\omega=\frac{\vic 1}{\vic \delta}(1-\epsilon)(1+e^{-1})(2+\epsilon)(1+\epsilon)^{\vic {\frac{\epsilon-1}{\epsilon}}}\ln\left(\frac{\vic 1}{\vic \delta}e^{\vic{\frac{1}{1-\epsilon}}}(1-\epsilon)(1+e^{-1})(2+\epsilon)(1+\epsilon)^{\vic {\frac{\epsilon-1}{\epsilon}}}\frac{\vic R^2}{\vic \gamma^2_{\rm d}}\right)$ and $0<\delta \le e^{-1}(1+e^{-1})(2+\epsilon)\frac{\vic R^2}{\vic \gamma^2_{\rm d}}$. For such a $b$ and taking into account the constraint on $\delta$ it can be verified that $\bar{t}_{\rm lb}\equiv \left((1+\epsilon)\frac{\vic b}{\vic \gamma^{\shri{1+\epsilon}}_{\rm d}}\right)^{\vic {\frac{1}{\epsilon}}}=(1+\epsilon)^{\vic {\frac{1}{\epsilon}}}\omega \frac{\vic R^2}{\vic \gamma^2_{\rm d}}$ satisfies the inequality $\bar{t}_{\rm lb}>e$. Moreover, any possible root of $\bar{g}(t)$ defined in (\ref{appr2}) and the single root $\vic {t_{{\rm b}_1}}$ of $g_{1}(t)$ are necessarily larger than $\bar{t}_{\rm lb}$. Therefore, since $\bar{t}_{\rm lb}> e$ and given that $\frac{\vic {d\bar{g}}}{\vic{dt}}> 0$ $\left(\frac{\vic d}{\vic {dt}}\frac{\vic {\ln t}}{\vic{t}}< 0\right)$ for $t>e$ there is a single root $\bar{t}_{\rm b}$ of $\bar{g}(t)$ satisfying $\bar{t}_{\rm b} \ge \vic {t_{{\rm b}_1}} > \bar{t}_{\rm lb}> e$. Combining (\ref{lfractb}) for $N=1$ with the last inequality and the relation $\bar{g}(\bar{t}_{\rm b})=0$ we get \vspace{-4pt}
\[
\begin{array}{l c}
f \ge \frac{\vic b}{\vic\gamma_{\rm d}^{\shri{1+\epsilon}}{\vic{t^{\epsilon}_{{\rm b}_1}}}} \ge \frac{\vic b}{\vic\gamma_{\rm d}^{\shri{1+\epsilon}}{\vic{\bar{\vic t}^{\epsilon}_{\rm b}}}} 
={\biggl\{\hspace{-2pt}\frac{\vic {R^2}}{\vic b\vic \gamma^{\shri{1-\epsilon}}_{\rm d}} \Bigl(\shh{\frac{1+\epsilon}{2}}\bar{t}_{\rm b}^{\epsilon}(\ln \bar{t}_{\rm b})^{1-\epsilon}+ \shh{\frac{3+\epsilon}{2}}\Bigr)\bar{t}_{\rm b}^{-1}+ 1+\epsilon \biggr\}^{-1}} & \\
>\inccc{{\biggl\{(2+\epsilon)\frac{\vic {R^2}}{\vic b\vic \gamma^{\shri{1-\epsilon}}_{\rm d}}
\Bigl(\frac{\vic{\ln \bar{t}_{\rm b}}}{\vic{\bar{t}_{\rm b}}}\Bigr)^{1-\epsilon}+ 1+\epsilon \biggr\}^{-1}}
> {\biggl\{(2+\epsilon)\frac{\vic {R^2}}{\vic b\vic \gamma^{\shri{1-\epsilon}}_{\rm d}}
\Bigl(\frac{\vic{\ln \bar{t}_{\rm lb}}}{\vic{\bar{t}_{\rm lb}}}\Bigr)^{1-\epsilon}+ 1+\epsilon \biggr\}^{-1}}}.  &
\end{array}
\] 
Let $x=\frac{\vic 1}{\vic \delta}e^{\vic{\frac{1}{1-\epsilon}}}(1-\epsilon)(1+e^{-1})(2+\epsilon)(1+\epsilon)^{\vic {\frac{\epsilon-1}{\epsilon}}}\frac{\vic R^2}{\vic \gamma^2_{\rm d}}$. Then, $\omega\frac{\vic R^2}{\vic \gamma^2_{\rm d}}=e^{-\vic{\frac{1}{1-\epsilon}}}x\ln x$ and
\[
\begin{array}{l}
(2+\epsilon)\frac{\vic {R^2}}{\vic b\vic \gamma^{\shri{1-\epsilon}}_{\rm d}}
\Bigl(\frac{\vic{\ln \bar{t}_{\rm lb}}}{\vic{\bar{t}_{\rm lb}}}\Bigr)^{1-\epsilon}\hspace{-2pt} =\hspace{-2pt}
(2+\epsilon)(1+\epsilon)^{\vic {\frac{\epsilon-1}{\epsilon}}}\omega^{-1}\hspace{-2pt}\left(\frac{\vic 1}{\vic \epsilon}\ln(1+\epsilon)+\ln\left(\omega\frac{\vic R^2}{\vic \gamma^2_{\rm d}}\right) \right)^{1-\epsilon}
\end{array}
\]
\begin{equation}
\label{eqremark3}
\begin{array}{l}
< \hspace{-2pt}(2+\epsilon)(1+\epsilon)^{\vic {\frac{\epsilon-1}{\epsilon}}}\omega^{-1}\hspace{-2pt}\left(1+(1-\epsilon)\ln\left(\omega\frac{\vic R^2}{\vic \gamma^2_{\rm d}}\right) \right)\hspace{-2pt}
=\hspace{-2pt} \frac{\vic{\delta}}{\vic {(1+e^{-1})}}\frac{\vic {\ln(x\ln x)}}{\vic{\ln x}} \le \delta
\end{array}
\end{equation}
($\ln\ln x/\ln x \le e^{-1}$). Thus, our choice of $b$ ensures that $f >(\delta+1+\epsilon)^{-1}$. Substituting $b$ into (\ref{lmartbound}) we conclude that in the $\ell$-margitron as $\epsilon, \delta \to 0$ the upper bound  on the number of updates $t_{\rm c}$ $\sim (\epsilon^{-1}+\delta^{-1}\ln\delta^{-1})\ln(\epsilon^{-1}+\delta^{-1}\ln\delta^{-1}){\vic R^2}/{\vic \gamma^2_{\rm d}}$. 
For $\epsilon \to 0$ with $\delta$ fixed, instead, the bound $\sim \epsilon^{-1}\ln\epsilon^{-1}{\vic R^2}/{\vic \gamma^2_{\rm d}}$.  For $\delta \ll 1$ and $\delta/\epsilon <\lambda \approx 1$, however, a more accurate upper bound on $t_{\rm c}$ may be obtained by observing that
\[
\begin{array}{l}
0=\bar{g}(\bar{t}_{\rm b})> {\vic{\bar{\vic t}^{\epsilon}_{\rm b}}}-(2+\epsilon)\frac{\vic {R^2}}{\vic\gamma^{2}_{\rm d}}\Bigl(\frac{\vic{\ln \bar{t}_{\rm b}}}{\vic{\bar{t}_{\rm b}}}\Bigr)^{1-\epsilon}-\vic{\bar{\vic t}^{\epsilon}_{\rm lb}}> {\vic{\bar{\vic t}^{\epsilon}_{\rm b}}}-(2+\epsilon)\frac{\vic {R^2}}{\vic\gamma^{2}_{\rm d}}\Bigl(\frac{\vic{\ln \bar{t}_{\rm lb}}}{\vic{\bar{t}_{\rm lb}}}\Bigr)^{1-\epsilon}-\vic{\bar{\vic t}^{\epsilon}_{\rm lb}}
\end{array}
\]
from where (using also (\ref{eqremark3}))
\[
\begin{array}{l}
\vic{\bar{\vic t}_{\rm b}}< \vic{\bar{\vic t}_{\rm lb}}\Bigl(1+(2+\epsilon)\frac{\vic{R^2}}{\vic\gamma^{2}_{\rm d}}{\vic{\bar{t}}^{-\epsilon}_{\rm lb}}\left(\frac{\vic{\ln \bar{t}_{\rm lb}}}{\vic{\bar{t}_{\rm lb}}}\right)^{1-\epsilon}\Bigr)^{\vic {\frac{1}{\epsilon}}}=\vic{\bar{\vic t}_{\rm lb}}\Bigl(1+\frac{\vic {2+\epsilon}}{\vic{1+\epsilon}}\;\frac{\vic {R^2}}{\vic b\vic \gamma^{\shri{1-\epsilon}}_{\rm d}}
\Bigl(\frac{\vic{\ln \bar{t}_{\rm lb}}}{\vic{\bar{t}_{\rm lb}}}\Bigr)^{1-\epsilon}\Bigr)^{\vic {\frac{1}{\epsilon}}}\\
<\vic{\bar{\vic t}_{\rm lb}}\left(1+\frac{\vic \delta}{\vic{1+\epsilon}}\right)^{\vic {\frac{1}{\epsilon}}}<{\bar{\vic t}_{\rm lb}}\left(1+\delta\right)^{\vic {\frac{1}{\epsilon}}}<\vic{\bar{\vic t}_{\rm lb}}e^{\vic {\frac{\delta}{\epsilon}}}=e^{\vic {\frac{\delta}{\epsilon}}}(1+\epsilon)^{\vic {\frac{1}{\epsilon}}}\omega \frac{\vic R^2}{\vic \gamma^2_{\rm d}}\enspace .
\end{array}
\]
Taking into account that $t_{\rm c}\le\vic {t_{{\rm b}_1}}\le\vic{\bar{\vic t}_{\rm b}}$ we conclude that as $\delta \to 0$ with ${\delta}/{\epsilon}$  bounded from above (e.g. $\delta=\epsilon \to 0$) the upper bound on $t_{\rm c}$ $\sim \delta^{-1}\ln \delta^{-1}{\vic R^2}/{\vic \gamma^2_{\rm d}}$.

\end{remark}

\begin{theorem}
\label{theorem4}
There is a value of the parameter $b$ for which the $\ell$-margitron with $\epsilon \ll 1$  converges to a solution hyperplane with directional margin $\gamma^\prime_{\rm d} \ge (1-2\epsilon)\gamma_{\rm d}$ in less than $\sim \epsilon^{-1}\ln \epsilon^{-1}{\vic R^2}/{\vic \gamma^2_{\rm d}}$ updates.
\end{theorem}

\begin{proof}
Set $\delta=\epsilon$ in Remark \ref{remark4} and notice that $f \ge (1+2\epsilon)^{-1} \ge 1-2\epsilon$. \medskip \hfill \fbox{}
\end{proof}

\begin{theorem}
\label{theorem5}
Both the t- and the $\ell$-margitron with $1< \epsilon <2$ converge in $t_{\rm c}$ updates, with $t_{\rm c}$ bounded from above by $\frac{\vic R^2}{\vic \gamma^2_{\rm d}}+\left(\shh{\frac{2}{2-\epsilon}}\frac{\vic b}{\vic \gamma^2_{\rm d}}\right)^{\vic {\frac{1}{\epsilon}}}$ and $\frac{\vic R^2}{\vic \gamma^2_{\rm d}}+\left(\shh{\frac{2}{2-\epsilon}}\frac{\vic b}{\vic \gamma^{\shri{1+\epsilon}}_{\rm d}}\right)^{\vic {\frac{1}{\epsilon}}}$ respectively, to a solution hyperplane possessing directional margin $\gamma^\prime_{\rm d}$ which in the limit $b \to \infty$ satisfies the inequality $\gamma^\prime_{\rm d} \ge (1-\frac{\epsilon}{2})\gamma_{\rm d}$.
\end{theorem}
\begin{proof}

For the t-margitron the analysis of Theorem \ref{theorem1} that led to (\ref{tineq}) remains valid and the single root $t_{\rm b}$ of $g(t)$ still provides an upper bound on $t_{\rm c}$. The bound on $t_{\rm c}$ stated in Theorem \ref{theorem5} is the upper bound on $t_{\rm b}$ inferred from Lemma \ref{lemma1}. The analysis that led to (\ref{tmarfrest}) remains also valid but we are no longer allowed to replace $t_{\rm c}$ with its lower bound $t_{\rm c}=1$. Instead, we may replace $t_{\rm c}$ in (\ref{tmarfrest}) with its upper bound stated in Theorem \ref{theorem5}. Then, as $b \to \infty$ we get $\gamma^\prime_{\rm d}\ge (1-\frac{\epsilon}{2})\gamma_{\rm d}$.

In the case of the $\ell$-margitron $\vec{a}_{t}\cdot\vec{y}_{k}$ for a misclassified pattern $\vec{y}_{k}$ may be bounded from above by employing (\ref{misclasl}) and (\ref{lbound}) as $\vec{a}_{t}\cdot\vec{y}_{k}\le b \left\|\vec a_t \right\|^{1-\epsilon}\le b(\gamma_{\rm d}t)^{1-\epsilon}$. Then, the analysis of Theorem \ref{theorem1}
that led to (\ref{tineq}) remains valid with the replacement of $b$ by $b\gamma_{\rm d}^{1-\epsilon}$. The bound on $t_{\rm c}$ stated in Theorem \ref{theorem5} is the upper bound on $t_{\rm b}$ inferred from Lemma \ref{lemma1}. For the fraction $f$, instead, employing (\ref{misclasl}), (\ref{nmisclas}), (\ref{tmarsq}) and (\ref{tineq}), with the last two relations taken at $t=t_{\rm b}$ as equalities, we have
\[
\begin{array}{c}
{f=\frac{\vic \gamma^\prime_{\rm d}}{\vic \gamma_{\rm d}}\ge  
\frac{\vic b}{\vic \gamma_{\rm d}\vic{\left\|\vec a_{t_{\rm c}} \right\|^{\epsilon}}}}
\ge \frac{\vic b}{\vic \gamma_{\rm d}^{\shri{1+\epsilon}}\vic{t_{\rm b}^{\epsilon}}}=
\left(\frac{\vic R^2}{\vic b \vic \gamma_{\rm d}^{\shri{1-\epsilon}}}t_{\rm b}^{\epsilon-1} + \shh{\frac{2}{2-\epsilon}}\right)^{-1}.
\end{array}
\]\vspace{-2pt}
Replacing $t_{\rm b}$ with its upper bound $\frac{\vic R^2}{\vic \gamma^2_{\rm d}}+\left(\shh{\frac{2}{2-\epsilon}}\frac{\vic b}{\vic \gamma^{\shri{1+\epsilon}}_{\rm d}}\right)^{\vic {\frac{1}{\epsilon}}}$ in the above relation leads to a weaker bound from where we get $\lim_{b \to \infty}f \ge 1-\frac{\epsilon}{2}$. \medskip \hfill \fbox{}
\end{proof}

\section{Experiments}
To reduce the computational cost we follow \cite{TST2} and form a reduced ``active set" of patterns consisting of the ones found misclassified during each epoch which are then cyclically presented to the Margitron algorithm for $N_{\rm ep}$ mini-epochs unless no update occurs during a mini-epoch. Subsequently, a new full epoch involving all the patterns takes place giving rise to a new active set. The algorithm terminates only if no mistake occurs during a full epoch. This procedure clearly amounts to a different way of sequentially presenting the patterns to the algorithm and does not affect the applicability of our theoretical analysis.

We compare the $t$- and the $\ell$-margitron with SVMs on the basis of their ability to achieve fast convergence to a certain approximation of the ``optimal" hyperplane in the feature space where the patterns are linearly separable. For linearly separable data the feature space is the initial instance space whereas for linearly inseparable data (which is the case here) a space extended by as many dimensions as the instances is considered where each instance is placed at a distance $\Delta$ from the origin in the corresponding dimension. The extension generates a margin of at least $\Delta/\sqrt{n}$ with $n$ being the number of patterns and amounts to adding a term $\Delta^2$ to the diagonal entries of the kernel (linear in our case). Moreover, its employment is justified by the well-known equivalence between the hard margin optimisation in the extended space and the soft margin optimisation in the initial instance space with objective function $\left\|{\vec w}\right\|^2+ \Delta^{-2} {\sum_i} {{\xi}_i}^2$ involving the weight vector $\vec w$ and the 2-norm of the slacks $\xi_i$ \cite{CST}.
We emphasize that SVMs and the Margitron are required to solve identical hard margin problems.

In our experiments SVMs are represented by ${\rm SVM}^{\rm light}$ \cite{Joa}, denoted here as ${\rm SVM}^{\rm l}$,  a decomposition method algorithm which is many orders of magnitude faster than standard SVMs. For ${\rm SVM}^{\rm l}$ we choose a memory parameter $m=400$MB and a 1-norm soft margin parameter $C=10^{5}$ (approximating $C=\infty$) since we are dealing with a hard margin problem in the appropriate feature space. The choice of the accuracy $\epsilon$ depends on the case. For the remaining parameters default values are used. The experiments were conducted on a 1.8 GHz Intel Pentium M processor with 504 MB RAM running Windows XP. The codes written in C++ were run using Microsoft's Visual C++ 5.0 compiler. 

The datasets we used for training are the Adult (32561 instances, 123 binary attributes) and Web (49749 instances, 300 binary attributes) UCI datasets as compiled by Platt (see \cite{Joa}), the test0 set from the Reuters RCV1 collection (199328 instances, 47236 attributes with average sparsity 0.16$\%$) obtainable from http://www.jmlr.org/papers /volume5/lewis04a/lyrl2004$\textunderscore$rcv1v2$\textunderscore$README.htm and the
multiclass Covertype (Cover) UCI dataset (581012 instances, 54 attributes).
In the case of the RCV1 we considered both the C11 and the CCAT binary text classification tasks while in the case of the Covertype dataset we studied the binary classification problem of the first class versus all the others. The Covertype dataset was rescaled by multiplying all the attributes with 0.001.

\begin{table}[t]
\centering
\caption{Results of a comparative study of ${\rm SVM}^{\rm l}$, $t$-margitron and  $\ell$-margitron.} 
\epsfig{file=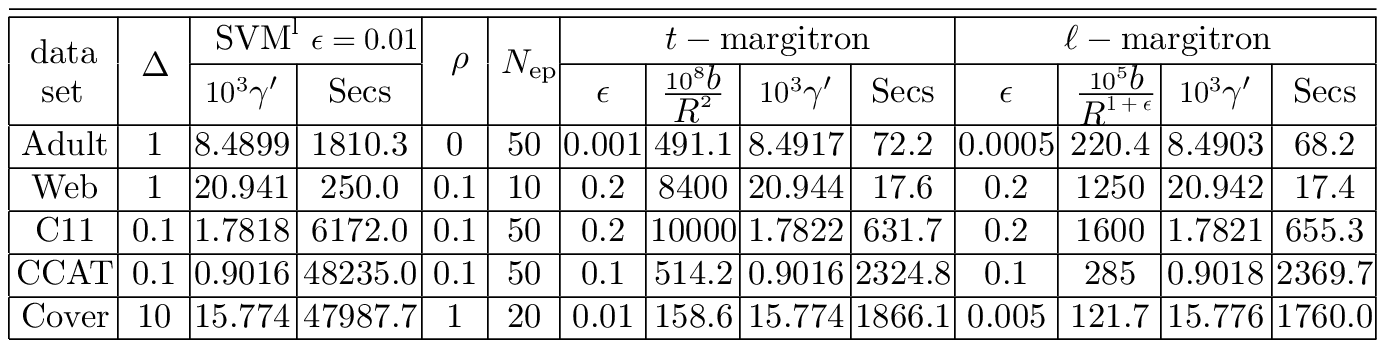, width=1\textwidth}
\end{table}

In Table 1 we present the results (i.e. geometric margin $\gamma^{\prime}$ achieved and CPU secs needed) of our first comparative study involving the algorithms ${\rm SVM}^{\rm l}$, $t$-margitron and  $\ell$-margitron together with the values of the parameters employed. A solution hyperplane in the extended space was first obtained using ${\rm SVM}^{\rm l}$ and subsequently the Margitron was required to obtain a solution of comparable geometric margin. The extended space parameter $\Delta$ refers to both ${\rm SVM}^{\rm l}$ and the Margitron while the augmented space parameter $\rho$ and the number of mini-epochs $N_{\rm ep}$ only to the Margitron. Also, for the Margitron $\gamma^{\prime}$ is the geometric margin in the original (non-augmented) feature space with the augmentation providing for the bias. We see that the Margitron is at least 10-20 times faster than ${\rm SVM}^{\rm l}$ on these rather large datasets. It is understood, of course, that some additional computer time was spent to locate the appropriate value of $b$. 

Recently ${\rm SVM}^{\rm perf}$ \cite{Joa06}, a cutting-plane algorithm for training linear SVMs, was presented. We did make an attempt at including ${\rm SVM}^{\rm perf}$ in our comparative study but we found that it requires a much longer CPU time to converge compared to ${\rm SVM}^{\rm l}$ without even achieving as large values of the margin $\gamma^{\prime}$. Table 2 contains our experimental results on the datasets Adult and Web ($\Delta=1$). Apparently, the ``accuracy" $\epsilon$ of ${\rm SVM}^{\rm perf}$ is not directly related to the fraction of the maximum margin achieved.

\begin{table}[hb]
\centering
\caption{Results of experiments with ${\rm SVM}^{\rm perf}$.}
\vspace{4pt}
\begin{tabular}{|@{\hspace{15pt}}c@{\hspace{15pt}}|@{\hspace{15pt}}c @{\hspace{15pt}}c @{\hspace{15pt}}c @{\hspace{15pt}}c@{\hspace{15pt}}|} 
\hline\hline
\multicolumn{1}{|c|@{\hspace{15pt}}}{\ \hspace{8pt} \mbox{\makebox(3,3)[t]{data}}  \hspace{8pt} \ }  & \multicolumn{4}{c|}{\  \hspace{8pt}${\rm SVM}^{\rm perf}$  \hspace{8pt} \  } \\[0.5pt]
\put(53,9){\line(1,0){180}}
\hspace{18pt} set & $\epsilon$ & $C$ & $\shr{10^3}\gamma^{\prime}$ & Secs   \\
\hline
\small{Adult} & $3\times10^{-4}$ & $10^8$ & 5.9436 & 54450.3 \\ 
\hline
\small{Web} & $2\times10^{-5}$ & $10^8$ & 20.891 & 7297.9 \\
\hline
\end{tabular}
\vspace{-10pt}
\end{table} 

In Table 3 we present the directional margin $\gamma^{\prime}_{\rm d}$ achieved by the t- and the $\ell$-margitron together with the after-running estimate $f_{\rm est}$ of the ratio ${\gamma^{\prime}_{\rm d}}/{\gamma_{\rm d}}$ and its asymptotic value for comparison. Let us accept that the geometric margin $\gamma^{\prime}$ reported in Table 1 is larger than $99 \%$ of the maximum geometric margin $\gamma$ as the accuracy $\epsilon=0.01$ of ${\rm SVM}^{\rm l}$ suggests. Then, taking into account that $\gamma \ge\gamma_{\rm d}$ and that $(\gamma^{\prime}-\gamma^{\prime}_{\rm d})/\gamma^{\prime}<0.02$ we see that ${\gamma^{\prime}_{\rm d}}/{\gamma_{\rm d}} > 0.97$. Thus, we may conclude that the estimates of Table 3 are certainly impressive given that they come from worst-case bounds which are not expected to be very tight and that they cannot, of course, exceed their asymptotic values.

\begin{table}[hb]
\centering
\caption{The directional margin $\gamma^{\prime}_{\rm d}$ achieved by the t- and the $\ell$-margitron together with the after-running estimate $f_{\rm est}$ of the ratio ${\gamma^{\prime}_{\rm d}}/{\gamma_{\rm d}}$ and its asymptotic value.}
\vspace{4pt}
\begin{tabular}{|@{\hspace{10pt}}c@{\hspace{10pt}} |@{\hspace{10pt}}c @{\hspace{13pt}}c @{\hspace{13pt}}c@{\hspace{10pt}}|@{\hspace{10pt}}c @{\hspace{13pt}}c @{\hspace{13pt}}c@{\hspace{10pt}}|} 
\hline\hline
\multicolumn{1}{|c|@{\hspace{10pt}}}{\ \mbox{\makebox(3,3)[t]{data}} \ } & \multicolumn{3}{c|@{\hspace{10pt}}}{\ $t-{\rm margitron}$ \  } &\multicolumn{3}{c|}{ \ $\ell-{\rm margitron}$ \ }\\[0.5pt]
\put(46.5,9){\line(1,0){252} }
\hspace{17pt}{set} & $\shr{10^3}\gamma^{\prime}_{\rm d}$ & $f_{\rm est}$ & $1\hspace{-3pt}-\hspace{-3pt}\frac{\epsilon}{2}$ & $\shr{10^3}\gamma^{\prime}_{\rm d}$ & $f_{\rm est}$ & \makebox{$\shhh{(1+\epsilon)^{-1}}$} \\
\hline
\small{Adult} & 8.4917 & 0.9898 & 0.9995 &  8.4903 & 0.9420 & 0.9995\\ 
\hline
\small{Web} & 20.574 & 0.8645 & 0.9000 & 20.573 & 0.7561 & 0.8333\\
\hline
\small{C11} & 1.7789 & 0.8923 & 0.9000 & 1.7787 & 0.8156 & 0.8333\\
\hline
\small{CCAT} & 0.9016 & 0.9404 & 0.9500 & 0.9018  &  0.8752 & 0.9091\\
\hline
\small{\makebox{Cover}} & 15.714 & 0.9873 & 0.9950 & 15.716 & 0.9703 & 0.9950\\
\hline
\end{tabular}
\end{table} 

From (\ref{bt}) and (\ref{bl}) it becomes apparent that the minimal value of $b$ guaranteeing the desired accuracy depends on the maximum directional margin $\gamma_{\rm d}$. Moreover, this dependence becomes increasingly crucial with decreasing $\epsilon$. This last observation prompts us to proceed to a determination of the large margin solution in successive runnings starting with the more insensitive to the value of $\gamma_{\rm d}$ Margitron with $\epsilon=1$ and gradually moving towards employing algorithms with smaller $\epsilon$'s able to guarantee larger fractions of $\gamma_{\rm d}$. Each running in this process will provide us with an interval in which the value of $\gamma_{\rm d}$ lies which, hopefully, will shrink as we move towards smaller $\epsilon$'s. This information will then allow us to fix the value of $b$ to be used in the next running. The lower bound on $\gamma_{\rm d}$ will be the margin $\gamma^{\prime}_{\rm d}$ achieved. The upper bound $\gamma^{\rm up}_{\rm d}$ will be provided by exploiting the after-running estimate $f_{\rm est}$ of $\gamma^{\prime}_{\rm d}/\gamma_{\rm d}$ which gives $\gamma^{\rm up}_{\rm d}=\gamma^{\prime}_{\rm d}f^{-1}_{\rm est}$. Alternatively, we may employ the upper bound on the number of updates $t_{\rm c}$ required for convergence to obtain a value for $\gamma^{\rm up}_{\rm d}$. For $\epsilon=1$ this gives $\gamma^{\rm up}_{\rm d}=R\sqrt{\left(1+2{\vic b}/{\vic R^2}\right)t^{-1}_{\rm c}}$ which is usually lower than the upper bound $\left({\vic R^2}/{\vic b}+2\right)\gamma^{\prime}_{\rm d}$ on $\gamma_{\rm d}$ obtained from $\gamma^{\prime}_{\rm d}/\gamma_{\rm d}\ge \left({\vic R^2}/{\vic b}+2\right)^{-1}$. This procedure may be followed using either the $t$- or the $\ell$-margitron but in the former case we may encounter difficulties for $\epsilon<1/2$ due to the lack of the strong before-running guarantees stemming from (\ref{tboundstr}). 

\begin{table}[t]
\centering
\caption{A comparison between the $\ell$-margitron (successive runnings) and ${\rm SVM}^l$.}
\begin{tabular}{|c|c|c@{\hspace{2pt}}|c|c@{\hspace{1pt}}|c@{\hspace{1pt}}|c|c|c|c|c|c@{\hspace{1pt}}|}
\hline\hline
\multicolumn{1}{|c|}{\  \ } & \multicolumn{8}{c|}{\ $\ell-{\rm margitron}$ \ } & \multicolumn{3}{c|}{\ \mbox{\makebox(12,2)[t]{${\rm SVM}^l$}} \ }\\
\cline{2-9}
\multicolumn{1}{|c|}{\ \mbox{\makebox(12,7)[t]{data}} \ } & \multicolumn{3}{c|}{\ $\epsilon=1$ \ } & \multicolumn{5}{c|}{\ $\epsilon=0.1$ \ } & \multicolumn{3}{c|}{\  \ }\\
\cline{2-12}
\mbox{\makebox(8,9)[t]{set}} & $\shhh{10^3} \gamma_{\rm d}^{\prime}$  & $\shhh{10^3}\gamma_{\rm d}^{\rm up}$ & Secs & $\frac{10^5\vic b}{\vic R^{1+\epsilon}}$ & $\shhh{10^3} \gamma_{\rm d}^{\prime}$ 
& $f_{\rm est}$ & $\shhh{10^3}\gamma^{\prime}$ & Secs & $\epsilon$ & $\shhh{10^3}\gamma^{\prime}$ & Secs \\
\hline 
\small{Adult} & 6.8839 & 11.352 & 3.2 & 577 & 8.3274 & 0.838 & 8.3274 & 39.5 &  0.055 & 8.3257 & 1178.9 \\
\hline 
\small{Web} & 19.202 & 29.840 & 5.0 & 551 & 20.677 & 0.864 & 21.053 & 44.0 & \makebox{0.0031} &  21.051 & 291.2 \\
\hline
\small{C11} & 1.5435 & 2.5607 & \makebox{121.7} & 506 & 1.7765 & 0.863 & 1.7798 & 774.7 & 0.012 &  1.7798 & 5952.0 \\
\hline
\small{CCAT} & 0.7800 & 1.2701 & \makebox{366.7} & 270 & 0.8989 & 0.855 & 0.8989 & \makebox{1771.5} & 0.0165 & 0.8984 & \makebox{42548.8} \\
\hline
\small{\makebox{Cover}} & 10.644 & 19.566 & \makebox{334.5} & 301 & 14.674 & 0.816 & 14.735 & 527.4 & 0.085 & 14.718 & \makebox{29402.8} \\
\hline

\end{tabular}
\end{table}

In Table 4 we present the results of a second comparative study between the $\ell$-margitron and ${\rm SVM}^{\rm l}$. For the $\ell$-margitron we followed the procedure of successive runnings that we just described involving only two stages with $\epsilon$ values 1 and 0.1. The extended and augmented feature spaces were identical to the ones of Table 1 and a common value $N_{\rm ep}=50$ was chosen for all datasets. Also, in the first stage ($\epsilon=1$) we made the common choice $b/R^2=5$ and obtained $\gamma^{\rm up}_{\rm d}$ from the relation $\gamma^{\rm up}_{\rm d}=R\sqrt{11t^{-1}_{\rm c}}$. Then, in the second stage ($\epsilon=0.1$) we fixed $b$ from (\ref{bl}) with $\delta=\left({\gamma_{\rm d}}/{\gamma^{\rm up}_{\rm d}}\right)^{{\vic {\frac{1-\epsilon}{\epsilon}}}}\le 1$ (which eliminates the dependence of $b$ on $\gamma_{\rm d}$) employing the $\gamma^{\rm up}_{\rm d}$ obtained in the first stage. This way we shift the uncertainty in ${\gamma_{\rm d}}/{\gamma^{\rm up}_{\rm d}}$ to the before-running accuracy $\delta$ and rely on the after-running lower bound $f_{\rm est}$ on $\gamma^{\prime}_{\rm d}/\gamma_{\rm d}$ to assess the accuracy actually achieved. We see that $f_{\rm est}$ is well above $0.8$ for all datasets.
A comparison with ${\rm SVM}^{\rm l}$ on solutions of comparable margin reveals that the $\ell$-margitron remains considerably faster even if the time spent to fix $b$ is taken into account.  

\section{Conclusions}
We generalised the classical Perceptron algorithm with margin by constructing the Margitron, a family of incremental large margin classifiers all the members of which employ the original perceptron update. The Margitron consists of two classes, namely the $t$-margitron with algorithms involving explicitly the number of updates and the $\ell$-margitron the members of which depend only on the length of the weight vector and as such lie closer in spirit to the Perceptron. We proved that as the parameter $\epsilon$ decreases from 2 to 0 the corresponding algorithms in both classes converge in a finite number of updates to hyperplanes possessing a guaranteed fraction of the maximum margin the largest possible value of which varies continuously in the interval $(0, 1)$. The Perceptron with margin belongs to both classes and is associated with the middle point of the above intervals. Finally, our experimental comparative study between algorithms from the margitron family and ${\rm SVM}^{\rm light}$ on tasks involving linear kernels and 2-norm soft margin revealed that the Margitron is a serious alternative to linear SVMs.

\end{document}